\documentclass[11pt]{article}

\usepackage{acl}

\usepackage{times}
\usepackage{latexsym}
\usepackage[T1]{fontenc}
\usepackage[utf8]{inputenc}
\usepackage{microtype}
\usepackage{inconsolata}
\usepackage{graphicx}

\usepackage{stfloats}
\usepackage[most]{tcolorbox}
\tcbuselibrary{listings,breakable,skins}
\usepackage{float}
\usepackage{booktabs}
\usepackage{multirow}
\usepackage{placeins}
\usepackage{xcolor}
\usepackage{amsmath,amssymb}
\usepackage{amsthm}
\usepackage{enumitem}

\theoremstyle{definition}
\newtheorem{definition}{Definition}
\newtheorem{problem}{Problem}

\newtcolorbox{casebox}[2][]{
  breakable,
  enhanced,
  width=\columnwidth,
  colback=gray!3!white,
  colframe=gray!65!black,
  colbacktitle=gray!65!black,
  coltitle=white,
  fonttitle=\bfseries\small,
  fontupper=\scriptsize\raggedright,
  title={#2},
  boxrule=0.4pt,
  arc=1mm,
  left=1.2mm,
  right=1.2mm,
  top=1.2mm,
  bottom=1.2mm,
  before skip=9pt,
  after skip=9pt,
  #1
}

\newtcblisting{jsoncasebox}[2][]{
  breakable,
  enhanced,
  listing only,
  width=\columnwidth,
  colback=gray!3!white,
  colframe=gray!65!black,
  colbacktitle=gray!65!black,
  coltitle=white,
  fonttitle=\bfseries\small,
  title={#2},
  boxrule=0.4pt,
  arc=1mm,
  left=1.2mm,
  right=1.2mm,
  top=1.2mm,
  bottom=1.2mm,
  before skip=9pt,
  after skip=9pt,
  listing options={
    basicstyle=\ttfamily\tiny,
    breaklines=true,
    breakatwhitespace=false,
    columns=fullflexible,
    keepspaces=true,
    showstringspaces=false,
    tabsize=2
  },
  #1
}

\title{From Rubrics to Reliable Scores: Evidence-Grounded Text Evaluation with LLM Judges}

\title{From Rubrics to Reliable Scores: Evidence-Grounded Text Evaluation with LLM Judges}

\author{
  \textbf{Yihan Hong\textsuperscript{1}} \quad
  \textbf{Huaiyuan Yao\textsuperscript{2}} \quad
  \textbf{Bolin Shen\textsuperscript{3}} \quad
  \textbf{Wanpeng Xu\textsuperscript{2}} \quad
  \textbf{Hua Wei\textsuperscript{2}} \quad
  \textbf{Yushun Dong\textsuperscript{3}}
  \\[2pt]
  \textsuperscript{1}Washington University in St. Louis, St. Louis, MO, USA \\
  \textsuperscript{2}Arizona State University, Tempe, AZ, USA \\
  \textsuperscript{3}Florida State University, Tallahassee, FL, USA
  \\[3pt]
  {\footnotesize\ttfamily
  h.yihan@wustl.edu \quad
  \{huaiyuan,wanpeng1,hua.wei\}@asu.edu \quad
  \{blshen,yushun.dong\}@fsu.edu
  }
}

\begin{document}

\maketitle

\begin{abstract}
Rubric-based text evaluation increasingly relies on large language models (LLMs) as scalable judges, yet frozen black-box models can interpret the same criteria inconsistently, produce score attributions that are difficult to audit, and map judgments poorly onto human scoring scales. We define this challenge as criteria transfer: translating human rubric intent into a stable, auditable inference-time scoring protocol. We introduce \textsc{Rulers}, which locks a task-level rubric specification, executes it through structured, evidence-grounded judgments, and calibrates the resulting signals to human score boundaries. Across four rubric-governed benchmarks and multiple frozen backbone models, \textsc{Rulers} achieves stronger agreement with human scores in most evaluated settings, while better matching empirical score distributions and remaining more stable under semantically equivalent rubric perturbations. Calibration controls and component ablations show that these gains cannot be attributed to post-hoc alignment alone, but depend on the combination of fixed criteria, traceable evidence, and calibrated score interpretation. These findings suggest that reliable LLM judging requires faithfully operationalizing human evaluation standards rather than relying on prompt-level scoring alone. Our code is available at \url{https://github.com/LabRAI/Rulers.git}.
\end{abstract}

\section{Introduction}

Rubric-based text evaluation is central to open-ended assessment tasks where outputs must be judged against human-defined criteria rather than surface-form overlap alone. Traditional reference-based metrics such as BLEU~\citep{papineni2002bleu} and ROUGE~\citep{lin2004rouge} provide scalable lexical signals, but they often miss higher-level qualities such as argument development, factual consistency, discourse organization, and instruction fulfillment~\citep{fabbri2021summeval,bhandari2020reevaluating}. This limitation is especially salient in student writing assessment and constructed-response scoring~\citep{mathias2018asapplusplus,crossley2022persuade,crossley2025asw_asap2}, as well as summarization evaluation~\citep{wang2020qags} and instruction-guided text generation~\citep{ye2024flask}. In these settings, evaluators must apply human-authored criteria to assess content, organization, relevance, factuality, and language quality. Large language model judges offer a scalable alternative for such tasks~\citep{zheng2023judging}, but scalable evaluation does not by itself ensure reliable scoring~\citep{gu2026survey}.

Reliable LLM-based scoring requires more than prompting a model to output a number. Although instruction-tuned LLMs can often follow complex instructions~\citep{ouyang2022training,chung2024scaling}, reliable evaluation requires them to apply criteria consistently, ground decisions in checkable evidence, and map their scoring tendencies to human score boundaries. We identify three recurring bottlenecks. First, rubric execution drift occurs when the same rubric is repeatedly reinterpreted at inference time, causing criteria definitions or score boundaries to vary with prompt phrasing, criterion order, output format, or backbone models. Such instability is consistent with prior findings that LLM judgments are sensitive to prompt design~\citep{zhao2021calibrate} and can exhibit position, verbosity, and self-enhancement biases~\citep{wang2024fair,zheng2023judging}. Second, unverifiable score attribution arises when scores or rationales are produced without evidence that can be checked against the input text. Free-form explanations may appear interpretable, but they are not necessarily faithful~\citep{turpin2023language} or sufficient for reliable evaluation~\citep{liu2023geval}. Third, human-scale misalignment occurs when raw model scores, trait judgments, or confidence signals do not match the score distributions and boundary conventions used by human raters, motivating post-hoc alignment and calibration~\citep{liu2024autocalibrate,gao2024bayesian}.

Our key observation is that a human rubric is not merely an instruction prompt, but a latent scoring structure. Rubrics specify quality dimensions, boundary descriptions, evidence expectations, and relationships between observable textual cues and final scores. We therefore formulate reliable rubric-based text evaluation as a criteria transfer problem: given a human-authored rubric, the goal is to transfer its scoring intent into a stable, auditable, and human-aligned protocol for a frozen LLM judge. This framing differs from generic preference ranking or broad open-ended LLM evaluation because we focus on rubric-governed text scoring settings where an explicit rubric and human reference scores define the target standard.

Prior work addresses parts of this problem through direct prompting and form-filling evaluation~\citep{liu2023geval}, specialized judge models~\citep{wang2024pandalm,zhu2025judgelm}, criteria decomposition~\citep{kim2024prometheus,liu2024hdeval}, and score calibration~\citep{liu2024autocalibrate,gao2024bayesian}. These approaches show that decomposition, grounding, and calibration are useful, but they are often treated as separate remedies rather than as a unified protocol for transferring human rubric intent into reliable scoring behavior.

We introduce \textsc{Rulers}, a three-stage framework for evidence-grounded criteria transfer in rubric-based text evaluation. \textsc{Rulers} first converts a human rubric into a structured and reusable specification, reducing runtime reinterpretation of criteria. It then executes the locked specification through an evidence-grounded scoring protocol that connects model judgments to extractive textual evidence and mechanically checks evidence validity when possible. Finally, it applies lightweight post-hoc calibration to align structured model signals with human score standards. Together, these stages address rubric execution drift, unverifiable score attribution, and human-scale misalignment without updating the judge model.

Our contributions are threefold. First, we formulate rubric-based text evaluation as a criteria-transfer problem, shifting the focus from whether LLMs can understand rubrics to whether human rubric intent can be transferred into stable, auditable, and human-aligned scoring procedures. Second, we propose \textsc{Rulers}, an inference-time framework that
effectively integrates rubric specification, evidence-grounded execution,
and human-scale calibration for frozen LLM judges. Third, we evaluate \textsc{Rulers} across multiple rubric-governed text evaluation settings, including essay scoring, summarization assessment, EFL writing assessment, and structured-input text generation, demonstrating improved human-score alignment and robustness across diverse datasets and backbone models.

\section{Preliminaries and Problem Formulation}
\label{sec:preliminaries}

\subsection{Notations}
\label{subsec:notations}

We consider rubric-based text scoring with a frozen LLM judge. Let $\mathcal{X}$ denote the space of evaluation instances, where each instance $x\in\mathcal{X}$ is a text response to be scored. Let $\mathcal{Y}=\{L,\ldots,U\}$ be the task-specific discrete human score scale, where $L$ and $U$ denote the minimum and maximum valid scores. Let $y\in\mathcal{Y}$ denote the human reference score.

Let $\mathcal{R}$ be the human-authored natural-language rubric for a task, and let $f_\theta$ be a frozen black-box LLM judge with fixed parameters $\theta$. We assume an annotated dataset $\mathcal{D}=\{(x_i,y_i)\}_{i=1}^{N}$, split into a labeled calibration set $\mathcal{D}_{\mathrm{cal}}$ and a held-out evaluation set $\mathcal{D}_{\mathrm{test}}$. The calibration set is used only to fit post-hoc score alignment, while $\mathcal{D}_{\mathrm{test}}$ is used only to report agreement with human scores.

A scoring protocol maps the raw rubric to a locked task-level rubric bundle $\mathcal{B}$ and applies this bundle to each input through the frozen judge. We denote the resulting structured scoring features by $\phi_\theta(x;\mathcal{B})$. A calibration function $g_\eta$ then maps these features to a final predicted score $\hat{y}\in\mathcal{Y}$. We use $\mathcal{A}(\cdot,\cdot)$ to denote a human-agreement metric between predicted and reference scores.

\subsection{Problem Formulation}
\label{subsec:formulation}

We focus on rubric-governed text scoring rather than general pairwise preference evaluation. In this setting, a human rubric and human reference scores define the target standard. We formalize the setting through the following constraints.

\begin{definition}[Frozen Rubric-Based Scoring Constraints]
\label{def:scoring_constraints}
\itshape
A valid scoring protocol in our setting must satisfy three constraints:
(i) frozen rubric-governed scoring, where $f_\theta$ is not updated and the target standard is specified by $\mathcal{R}$ and $\mathcal{Y}$;
(ii) fixed task-level execution, where $\mathcal{R}$ is transformed into a reusable bundle $\mathcal{B}$, the same bundle is used for all instances in the task, and cited evidence is checked against the input text; and
(iii) calibration-only human-scale alignment, where the mapping from structured model signals to final scores is fitted only on $\mathcal{D}_{\mathrm{cal}}$ and then fixed for held-out scoring.
\end{definition}

Under Definition~\ref{def:scoring_constraints}, the evaluator cannot improve human agreement by fine-tuning the judge, rewriting the rubric for each instance, relying on unconstrained free-form rationales, or adapting to held-out labels. Let $\Omega(\mathcal{R},\mathcal{D}_{\mathrm{cal}})$ denote the set of valid inference-time scoring protocols satisfying these constraints.

\begin{problem}[Human-Aligned Rubric-Based Text Scoring]
\label{prob:human_aligned_scoring}
\itshape
Given a human rubric $\mathcal{R}$, a frozen LLM judge $f_\theta$, a labeled calibration set $\mathcal{D}_{\mathrm{cal}}$, and the constraints in Definition~\ref{def:scoring_constraints}, the objective is to construct an inference-time scoring protocol $\mathcal{P}$ that predicts a final score $\hat{y}^{\mathcal{P}}(x)\in\mathcal{Y}$ for each held-out instance $x$. We write
\begin{equation}
\hat{y}^{\mathcal{P}}(x)
=
g_{\eta^\star}\!\left(
\phi_\theta(x;\pi(\mathcal{R}))
\right),
\label{eq:protocol_prediction}
\end{equation}
where $\pi(\mathcal{R})=\mathcal{B}$ is the rubric specification step, $\phi_\theta$ denotes the structured scoring features obtained from the frozen judge and locked bundle, and $g_{\eta^\star}$ is fitted only on $\mathcal{D}_{\mathrm{cal}}$.

The criteria-transfer objective is to maximize agreement with human scores
while satisfying stability and auditability constraints. Let
$\mathcal{A}_{\mathrm{test}}(\mathcal{P})$ denote the human-agreement metric
$\mathcal{A}$ evaluated between the predicted and reference scores on
$\mathcal{D}_{\mathrm{test}}$. We then write
\begin{equation}
\begin{aligned}
\mathcal{P}^{\star}
&=
\arg\max_{\mathcal{P}\in
\Omega(\mathcal{R},\mathcal{D}_{\mathrm{cal}})}
\mathcal{A}_{\mathrm{test}}(\mathcal{P}) \\
\text{s.t.}\qquad
\mathrm{Instab}(\mathcal{P})
&\leq \epsilon_s, \\
\mathrm{Audit}(\mathcal{P})
&\geq \gamma .
\end{aligned}
\label{eq:human_aligned_scoring_goal}
\end{equation}

Here, $\mathrm{Instab}(\mathcal{P})$ measures rubric-execution instability,
namely the expected prediction disagreement when the same source rubric is
semantically reworded, reordered. $\mathrm{Audit}(\mathcal{P})$ measures evidence-grounding coverage,
namely the fraction of returned evidence items that satisfy their
corresponding verification rules. The tolerances $\epsilon_s$ and $\gamma$
specify the desired upper bound on instability and lower bound on
auditability, respectively.
\end{problem}

\section{Methodology}
\label{sec:methodology}

\subsection{Overview}
\label{subsec:method_overview}

\begin{figure*}[t]
    \centering
    \includegraphics[width=0.85\textwidth]{{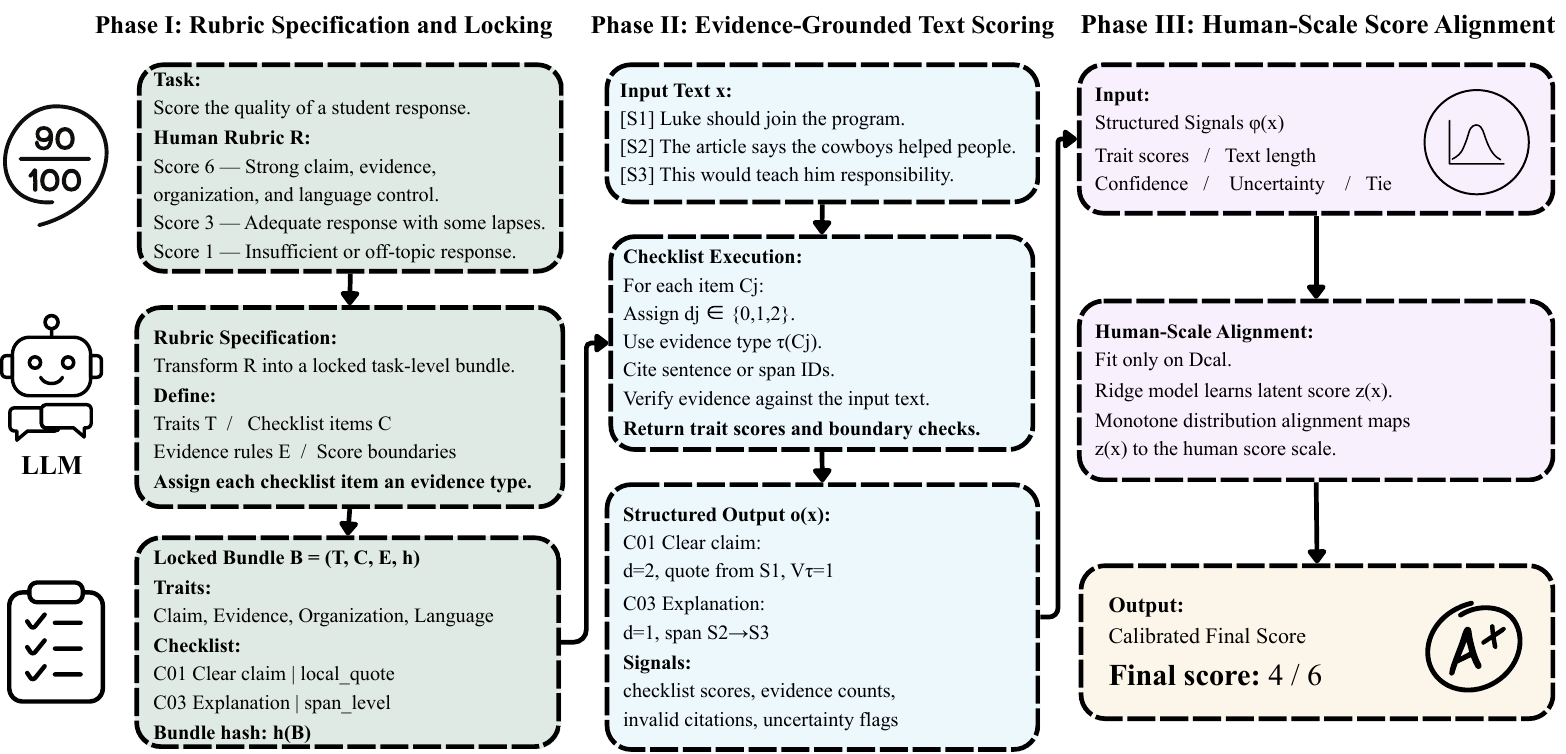}}
    \caption{Overview of the \textsc{Rulers} pipeline for rubric-based text scoring.}
    \label{fig:rulers_framework}
    \vspace{-8pt}
\end{figure*}

\textsc{Rulers} instantiates the scoring protocol in Problem~\ref{prob:human_aligned_scoring}. It separates rubric-based scoring into rubric specification, evidence-grounded execution, and human-scale calibration. This separation fixes rubric interpretation at the task level, makes score attribution auditable, and aligns structured model signals to human score boundaries. Table~\ref{tab:rulers_pipeline} summarizes the pipeline, Figure~\ref{fig:rulers_framework} gives the overview of the pipeline, and Appendix~\ref{app:case_study} provides a concrete ASAP 2.0 case sample.

\begin{table}[t]
\centering
\small
\setlength{\tabcolsep}{3pt}
\caption{Pipeline summary of \textsc{Rulers}.}
\label{tab:rulers_pipeline}
\resizebox{\columnwidth}{!}{
\begin{tabular}{llll}
\toprule
Stage & Input & Operation & Output \\
\midrule
I & Rubric $\mathcal{R}$ & Specify and lock & Bundle $\mathcal{B}$ \\
II & Text $x$, bundle $\mathcal{B}$ & Execute and verify & Signals $\phi_\theta(x;\mathcal{B})$ \\
III & Signals, $\mathcal{D}_{\mathrm{cal}}$ & Calibrate & Score $\hat{y}$ \\
\bottomrule
\end{tabular}
}
\end{table}

\subsection{Phase I: Rubric Specification and Locking}
\label{subsec:phase1}
\label{sec:phase1}

Phase I reduces runtime reinterpretation by inducing a structured rubric bundle from the rubric $\mathcal{R}$:
\begin{equation}
\mathcal{B}=(\mathcal{T},\mathcal{C},\mathcal{S},\mathcal{E},h),
\label{eq:rubric_bundle}
\end{equation}
where $\mathcal{T}$ denotes evaluation traits, $\mathcal{C}$ is an operational checklist, $\mathcal{S}$ contains task-specific score anchors and boundary descriptions, $\mathcal{E}$ specifies evidence rules, and $h$ is the hash of the serialized bundle.

Each checklist item $c_j\in\mathcal{C}$ is assigned to a trait $t(c_j)\in\mathcal{T}$ and an evidence-grounding type $\tau(c_j)$, and receives a decision $d_j\in\{0,1,2\}$, denoting absence, partial presence, and clear presence, respectively. The evidence rules $\mathcal{E}$ define how cited evidence should be anchored to the input text for each checklist type. Thus, different checklist items may require different forms of evidence depending on the nature of the criterion; the detailed type-to-operator mapping is provided in Appendix~\ref{app:evidence_operator_types}. Although bundle induction uses LLM generation and is not assumed to be deterministic, stability comes from locking: once induced, $\mathcal{B}$ is serialized, hashed, and reused unchanged for all instances in the same task. A concrete Phase I example is provided in Appendix~\ref{app:case_phase1}.

\subsection{Phase II: Evidence-Grounded Rubric Execution}
\label{subsec:phase2}
\label{sec:phase2}

Given an input text $x$, \textsc{Rulers} segments it into atomic text units $\mathcal{U}_x=\{u_1,\ldots,u_M\}$, such as sentences or paragraph-level units. The frozen judge receives $\mathcal{B}$ and $\mathcal{U}_x$, and returns a structured output $o(x)=(\mathbf{d},E_x,r(x))$, where $\mathbf{d}$ is the checklist decision vector, $E_x$ contains cited evidence, and $r(x)$ contains auxiliary verification or uncertainty indicators.

We aggregate checklist decisions into raw trait scores for calibration. 
Let $\mathcal{C}_k$ be the checklist items assigned to trait $t_k$. We first compute
\begin{equation}
\mu_k=\frac{\sum_{c_j\in\mathcal{C}_k}d_j}{2|\mathcal{C}_k|}.
\label{eq:trait_score_compact}
\end{equation}
The raw trait score $s_k$ is obtained by rounding and clipping the linear mapping of $\mu_k$ to the trait range $\mathcal{Y}_k=\{L_k,\ldots,U_k\}$.

To make attribution auditable, \textsc{Rulers} requires evidence to follow the operator specified by the checklist type $\tau(c_j)$. We verify grounding by setting $V_{\tau}(e,\mathcal{U}_x)=1$ when the returned evidence $e$ satisfies the corresponding operator-specific rule, and 0 otherwise. This allows the same locked checklist to support different evidence requirements while preserving a unified structured output for downstream calibration. Checklist decisions, trait scores, and uncertainty indicators are then passed to Phase III as structured scoring signals. A concrete Phase II example is provided in Appendix~\ref{app:case_phase2}.

\subsection{Phase III: Human-Scale Distribution Alignment}
\label{subsec:phase3}
\label{sec:phase3}

The structured signals from Phase II are not assumed to match human score boundaries directly: a frozen LLM may be systematically stricter or more lenient than human raters, and tasks may have different score distributions. Phase III therefore aligns these signals to the human scale using only the labeled calibration set $\mathcal{D}_{\mathrm{cal}}$.

For each instance, we construct $\phi_\theta(x;\mathcal{B})$ from trait scores $\{s_k\}_{k=1}^{K}$ and auxiliary signals, then fit a second-order ridge regression model on $\mathcal{D}_{\mathrm{cal}}$ to obtain a latent score:
\begin{equation}
z(x)=
\mathrm{Ridge}_{\alpha}
\left(
\mathrm{Std}
\left(
\psi_2(\phi_\theta(x;\mathcal{B}))
\right)
\right),
\label{eq:ridge_latent_score}
\end{equation}
where $\psi_2(\cdot)$ denotes second-order polynomial features and $\mathrm{Std}(\cdot)$ denotes feature standardization fitted on the calibration split.

The latent score is mapped to the human score scale through monotone distribution alignment:
\begin{equation}
\hat{y}
=
\mathrm{Clip}_{\mathcal{Y}}
\left(
\mathrm{Round}
\left[
\widehat{F}^{-1}_{Y}
\left(
\widehat{F}_{Z}(z(x))
\right)
\right]
\right),
\label{eq:distribution_alignment}
\end{equation}
where $\widehat{F}_{Z}$ and $\widehat{F}_{Y}$ are the empirical cumulative distribution functions of calibration latent scores and human scores, respectively. In practice, we implement this mapping by monotone interpolation on the calibration split. After fitting, both the ridge model and distribution map are fixed and applied unchanged to test instances. A concrete Phase III example is provided in Appendix~\ref{app:case_phase3}.

\section{Experiments}
\label{sec:experiments}

We evaluate \textsc{Rulers} across rubric-governed text scoring tasks with diverse score scales, rubric structures, and task domains.

\subsection{Experimental Setup}
\label{subsec:exp_setup}

\paragraph{Datasets.}
We evaluate on four rubric-governed benchmarks spanning different text evaluation settings and human-score constructions. ASAP 2.0~\citep{crossley2025asw_asap2} is a source-based argumentative essay scoring dataset with direct holistic writing-quality scores. SummHF~\citep{stiennon2020learning} evaluates summarization quality from human feedback and provides direct overall human scores. DREsS~\citep{yoo2025dress} is a large-scale rubric-based EFL essay scoring dataset whose official overall score aggregates Content, Organization, and Language ratings. WebNLG~\citep{gardent2017webnlg} evaluates structured-input text generation from RDF triples; because it provides separate ratings rather than a native overall score, we use their sum as the task-level target. These datasets also vary in score range, rubric granularity, evidence requirements, and level of subjectivity. Further statistics and preprocessing details are provided in Appendix~\ref{sec:appendix_datasets}.

\paragraph{Backbone Models.}
We test both proprietary and open-weight frozen judges without task-specific fine-tuning: GPT-4o, GPT-4o-mini, Llama-3.1-8B-Instruct, and Llama-3.1-70B-Instruct. Within each comparison, all methods use the same backbone, input instance, and decoding setting. All inference is conducted with temperature $0.0$.

\paragraph{Baselines.}
We compare \textsc{Rulers} with four inference-time baselines. Direct Holistic Scoring (DHS) prompts the model to apply the raw rubric and output one final score~\citep{zheng2023judging,liu2023geval}. Multi-Trait Specialization (MTS) scores rubric traits separately and aggregates them~\citep{lee2024unleashing}. AutoScore extracts rubric-relevant components before assigning a score~\citep{wang2026autoscore}. CheckEval converts criteria into checklist-style questions and aggregates checklist responses~\citep{lee2025checkeval}. These baselines follow their original inference-time protocols and do not use the post-hoc calibration layer introduced in \textsc{Rulers}. Appendix~\ref{app:baseline_protocol} provides a compact protocol-level comparison.

\paragraph{Calibration Protocol.}
We use $N=200$ labeled calibration examples and never use held-out labels for fitting. The alignment model uses second-order polynomial features, standardization, ridge regression with $\alpha=2.5$, and monotone quantile mapping to the empirical human score distribution. Section~\ref{subsec:calibration_control} varies $N\in\{50,100,150,200\}$ and applies the same procedure to baselines.

\paragraph{Evaluation Metrics.}
We use Quadratic Weighted Kappa (QWK)~\citep{cohen1968weighted} as the primary metric, following automated scoring practice~\citep{shermis2012contrasting}. QWK measures chance-corrected agreement and penalizes larger score disagreements more heavily:
\begin{equation}
\mathrm{QWK}
=
1-
\frac{\sum_{i,j\in\mathcal{Y}} w_{ij}O_{ij}}
{\sum_{i,j\in\mathcal{Y}} w_{ij}Q_{ij}},
\quad
w_{ij}=\frac{(i-j)^2}{(U-L)^2}.
\label{eq:qwk}
\end{equation}
Here $O$ is the observed agreement matrix, $Q$ is the expected agreement matrix under empirical score marginals, and $L$ and $U$ are the minimum and maximum scores in the task-specific scale. Higher QWK indicates stronger agreement with human scores.

\subsection{Quantitative Performance on Human Alignment}
\label{subsec:main_results}

We evaluate human-score alignment across four rubric-governed benchmarks and four backbone models. Table~\ref{tab:main_results_wide} reports QWK with bootstrap standard errors computed over held-out evaluation instances. Higher values indicate stronger agreement with human scores. To contextualize the inference cost of each method, Appendix~\ref{app:token_usage} further reports the average token usage per evaluated sample.

\providecommand{\tightpm}{\!\pm\!}

\begin{table*}[t]
\centering
\scriptsize
\setlength{\tabcolsep}{3.2pt}
\renewcommand{\arraystretch}{1.08}
\caption{
Performance comparison in QWK. Values are reported as mean$\pm$bootstrap SE. Best values in each row are highlighted in bold.
}
\label{tab:main_results_wide}
\resizebox{\textwidth}{!}{
\begin{tabular}{@{}llccccc@{}}
\toprule
\textbf{Dataset} & \textbf{Backbone} 
& \textbf{DHS} 
& \textbf{MTS} 
& \textbf{AutoScore} 
& \textbf{CheckEval} 
& \textbf{\textsc{Rulers}} \\
\midrule

\multirow{4}{*}{ASAP 2.0}
& 4o-mini 
& $0.4584\tightpm0.0079$ 
& $0.5566\tightpm0.0082$ 
& $0.4653\tightpm0.0083$ 
& $0.4629\tightpm0.0077$ 
& $\mathbf{0.7077\tightpm0.0058}$ \\

& GPT-4o 
& $0.4433\tightpm0.0076$ 
& $0.4939\tightpm0.0087$ 
& $0.3912\tightpm0.0085$ 
& $0.4466\tightpm0.0081$ 
& $\mathbf{0.7122\tightpm0.0056}$ \\

& L3.1-8B 
& $0.1973\tightpm0.0049$ 
& $0.4227\tightpm0.0097$ 
& $0.3548\tightpm0.0087$ 
& $0.3169\tightpm0.0078$ 
& $\mathbf{0.6827\tightpm0.0060}$ \\

& L3.1-70B 
& $0.4804\tightpm0.0079$ 
& $0.4938\tightpm0.0090$ 
& $0.4591\tightpm0.0086$ 
& $0.3796\tightpm0.0083$ 
& $\mathbf{0.7179\tightpm0.0055}$ \\

\midrule

\multirow{4}{*}{SummHF}
& 4o-mini 
& $0.3275\tightpm0.0238$ 
& $0.0886\tightpm0.0269$ 
& $0.0660\tightpm0.0174$ 
& $0.2739\tightpm0.0211$ 
& $\mathbf{0.3984\tightpm0.0305}$ \\

& GPT-4o 
& $0.1665\tightpm0.0140$ 
& $0.0884\tightpm0.0267$ 
& $0.3358\tightpm0.0179$ 
& $0.3278\tightpm0.0203$ 
& $\mathbf{0.4153\tightpm0.0313}$ \\

& L3.1-8B 
& $0.1205\tightpm0.0209$ 
& $0.1625\tightpm0.0226$ 
& $0.0419\tightpm0.0235$ 
& $\mathbf{0.2594\tightpm0.0276}$ 
& $0.2060\tightpm0.0318$ \\

& L3.1-70B 
& $0.4258\tightpm0.0226$ 
& $0.1625\tightpm0.0220$ 
& $0.0419\tightpm0.0232$ 
& $\mathbf{0.4606\tightpm0.0256}$ 
& $0.4454\tightpm0.0315$ \\

\midrule

\multirow{4}{*}{DREsS}
& 4o-mini 
& $0.2395\tightpm0.0099$ 
& $0.2935\tightpm0.0195$ 
& $0.1961\tightpm0.0182$ 
& $0.1052\tightpm0.0065$ 
& $\mathbf{0.5292\tightpm0.0203}$ \\

& GPT-4o 
& $0.2836\tightpm0.0165$ 
& $0.3181\tightpm0.0190$ 
& $0.1557\tightpm0.0155$ 
& $0.2062\tightpm0.0171$ 
& $\mathbf{0.5369\tightpm0.0202}$ \\

& L3.1-8B 
& $0.2489\tightpm0.0167$ 
& $0.2024\tightpm0.0171$ 
& $0.0446\tightpm0.0094$ 
& $0.2236\tightpm0.0183$ 
& $\mathbf{0.5574\tightpm0.0195}$ \\

& L3.1-70B 
& $0.2965\tightpm0.0167$ 
& $0.3712\tightpm0.0211$ 
& $0.1711\tightpm0.0177$ 
& $0.1142\tightpm0.0104$ 
& $\mathbf{0.5369\tightpm0.0205}$ \\

\midrule

\multirow{4}{*}{WebNLG}
& 4o-mini 
& $0.4924\tightpm0.0222$ 
& $0.5718\tightpm0.0259$ 
& $0.5701\tightpm0.0246$ 
& $0.5624\tightpm0.0262$ 
& $\mathbf{0.6135\tightpm0.0294}$ \\

& GPT-4o 
& $0.5001\tightpm0.0229$ 
& $0.5019\tightpm0.0236$ 
& $0.4062\tightpm0.0217$ 
& $0.6187\tightpm0.0223$ 
& $\mathbf{0.6274\tightpm0.0232}$ \\

& L3.1-8B 
& $0.4150\tightpm0.0231$ 
& $0.4474\tightpm0.0247$ 
& $0.3808\tightpm0.0218$ 
& $0.5229\tightpm0.0306$ 
& $\mathbf{0.5899\tightpm0.0264}$ \\

& L3.1-70B 
& $0.6450\tightpm0.0195$ 
& $0.4897\tightpm0.0229$ 
& $0.6367\tightpm0.0223$ 
& $0.6337\tightpm0.0245$ 
& $\mathbf{0.6495\tightpm0.0228}$ \\

\bottomrule
\end{tabular}
}
\vspace{-4pt}
\end{table*}

Overall, \textsc{Rulers} achieves stronger human-score alignment in most evaluated settings and remains relatively stable across backbone models. The improvement is not uniform, however: on SummHF, \textsc{Rulers} is not the best-performing method with the Llama-based judges. This may reflect backbone-specific differences in reliably executing the structured rubric and producing informative intermediate signals for this task. This pattern suggests that the gains of \textsc{Rulers} are not tied to a single proprietary or open-weight judge, but instead arise from the combination of locked rubric execution, evidence-grounded scoring signals, and human-scale calibration. The results also show that stronger backbone models do not always lead to higher QWK, highlighting that rubric-based evaluation depends not only on model capability but also on score-boundary alignment. Although \textsc{Rulers} uses more tokens than lightweight single-pass methods because it returns structured checklist decisions, evidence fields, and scoring signals, its average token usage remains lower than MTS, which requires multiple trait-level LLM calls per sample. These findings support our central claim that reliable rubric-based scoring requires controlling both rubric execution and the mapping from model-derived signals to human scoring standards.

\subsection{Distribution Alignment and Stability}
\label{subsec:dist_align}

\begin{figure*}[t]
\centering
\includegraphics[width=\textwidth]{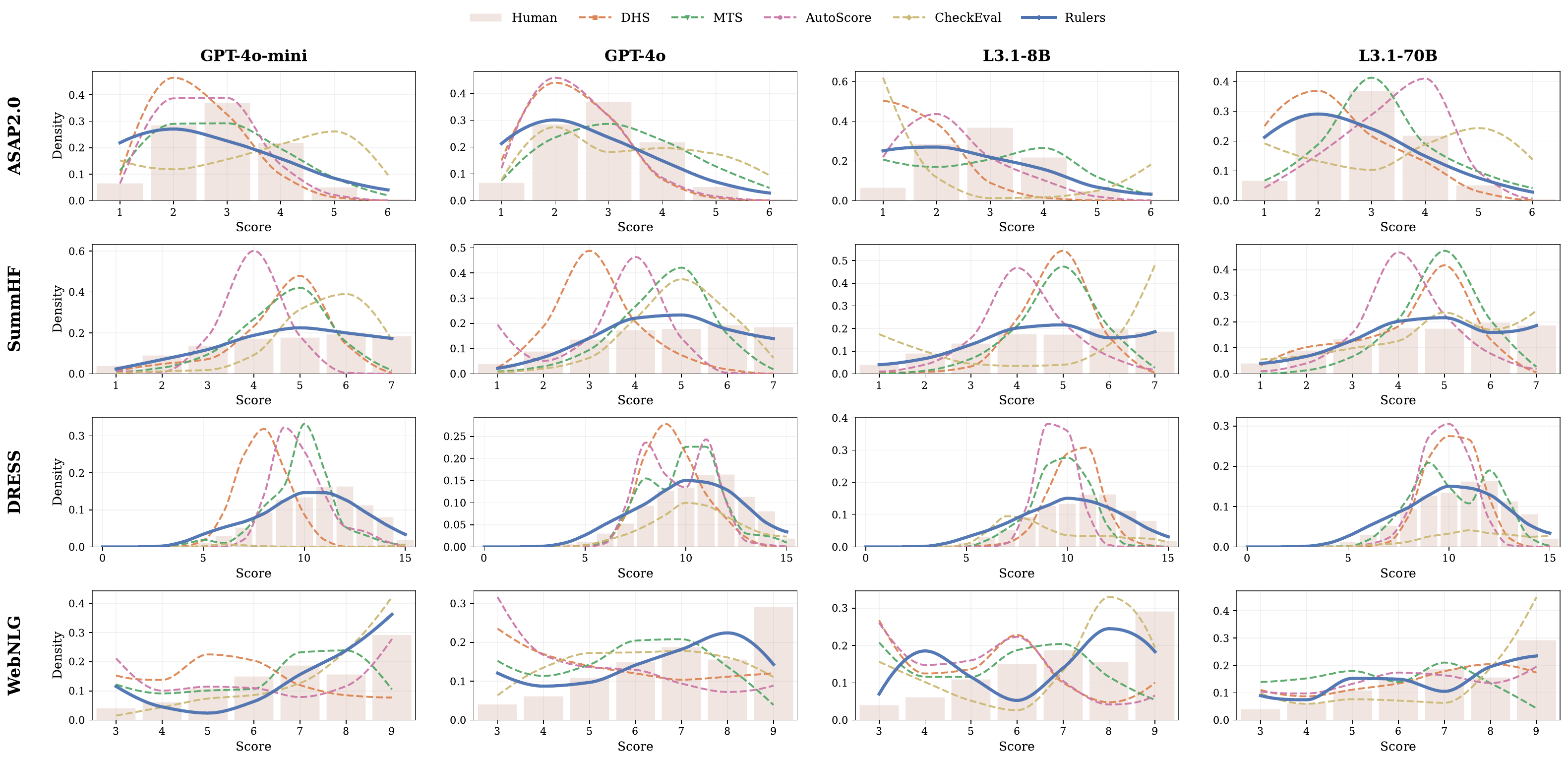}
\caption{
Score distribution alignment across datasets and backbone models. Pink histograms show the empirical human score distributions, and colored curves show the predicted score distributions of different methods.
}
\label{fig:dist_align}
\end{figure*}

Beyond point-wise agreement, we examine whether each method produces score distributions that align with human ratings. Figure~\ref{fig:dist_align} compares predicted score distributions with the empirical human distribution across four benchmarks and four backbone models. The human scores are shown as histograms, while each method is shown as a smoothed density curve.

\textsc{Rulers} shows stronger distributional alignment with human ratings across most settings. Compared with methods that directly score from raw rubrics or aggregate unconstrained intermediate outputs, its predicted distributions more often preserve both the central mass and the spread of human scores. The patterns are also more stable across backbone models, suggesting that locked rubric execution and post-hoc human-scale calibration reduce sensitivity to model-specific score priors. These patterns complement the QWK scores in Table~\ref{tab:main_results_wide}, suggesting that the gains reflect not only point-wise score agreement but also improved use of the human rating scale.

\subsection{Calibration Control and Data Efficiency}
\label{subsec:calibration_control}

Because \textsc{Rulers} includes a Phase III distribution-alignment step, we further test whether its gains can be explained by calibration alone. Using \texttt{GPT-4o-mini}, we apply the same second-order ridge plus monotone distribution-alignment procedure to the final predicted scores of baseline methods, fitting only on the calibration split and evaluating on held-out samples. Table~\ref{tab:calib_control_compact} reports a compact comparison for DHS, CheckEval, and \textsc{Rulers} at $N=100$ and $N=200$ calibration examples. Full results for all methods and $N\in\{50,100,150,200\}$ are provided in Appendix~\ref{app:calib_control_full}. We further compare the QWK gain from applying the same post-hoc calibration procedure to DHS with the remaining gap between the calibrated DHS baseline and \textsc{Rulers}. This comparison helps determine how much of the improvement comes from calibration alone versus the structured scoring stages of \textsc{Rulers}, and is reported in Appendix~\ref{app:calibration_gain_diagnostic}.

\begin{table}[t]
\centering
\scriptsize
\setlength{\tabcolsep}{3.5pt}
\renewcommand{\arraystretch}{1.08}
\caption{
Compact calibration-control comparison. Values are QWK mean$\pm$bootstrap SE.
}
\label{tab:calib_control_compact}
\resizebox{\columnwidth}{!}{
\begin{tabular}{llcc}
\toprule
\textbf{Dataset}
& \textbf{Method}
& \textbf{$N=100$}
& \textbf{$N=200$} \\
\midrule

\multirow{3}{*}{ASAP 2.0}
& DHS
& $0.5414\tightpm0.0080$
& $0.5413\tightpm0.0081$ \\
& CheckEval
& $0.5450\tightpm0.0082$
& $0.5457\tightpm0.0082$ \\
& \textsc{Rulers}
& $\mathbf{0.7024\tightpm0.0058}$
& $\mathbf{0.7077\tightpm0.0058}$ \\

\midrule

\multirow{3}{*}{SummHF}
& DHS
& $\mathbf{0.3743\tightpm0.0254}$
& $0.3655\tightpm0.0260$ \\
& CheckEval
& $0.3683\tightpm0.0269$
& $0.3891\tightpm0.0269$ \\
& \textsc{Rulers}
& $0.3388\tightpm0.0308$
& $\mathbf{0.3984\tightpm0.0305}$ \\

\midrule

\multirow{3}{*}{DREsS}
& DHS
& $0.3439\tightpm0.0215$
& $0.3448\tightpm0.0216$ \\
& CheckEval
& $0.3293\tightpm0.0219$
& $0.3284\tightpm0.0217$ \\
& \textsc{Rulers}
& $\mathbf{0.4977\tightpm0.0204}$
& $\mathbf{0.5292\tightpm0.0203}$ \\

\midrule

\multirow{3}{*}{WebNLG}
& DHS
& $0.5605\tightpm0.0211$
& $0.5608\tightpm0.0209$ \\
& CheckEval
& $0.5880\tightpm0.0251$
& $0.5822\tightpm0.0250$ \\
& \textsc{Rulers}
& $\mathbf{0.5997\tightpm0.0251}$
& $\mathbf{0.6135\tightpm0.0294}$ \\

\bottomrule
\end{tabular}
}
\vspace{-4pt}
\end{table}

The calibration-control results show that distribution alignment can improve QWK, but the quality of the scores before alignment remains crucial. Applying the same alignment procedure to baseline outputs does not consistently reproduce the gains of \textsc{Rulers}, indicating that calibration is most effective when the upstream scoring signals are already structured, rubric-grounded, and informative. The calibration-size comparison also suggests that $N=100$ labeled examples already yields strong performance in many settings, while $N=200$ provides additional but generally moderate gains. This supports the interpretation that Phase III helps align score scales, but the advantage of \textsc{Rulers} also depends on the preceding rubric specification and evidence-grounded scoring stages.

\subsection{Rubric Sensitivity Under Perturbations}
\label{subsec:robustness}

We first verify that Phase-I rubric transfer preserves the intended
source-rubric structure; detailed fidelity results are provided in
Appendix~\ref{app:rubric_transfer_fidelity}. To evaluate sensitivity to rubric presentation, we test each method under three semantically equivalent rubric variants using the same \texttt{GPT-4o-mini} backbone: the original rubric (\textit{Standard}), a criterion-order-inverted rubric (\textit{Reversed}), and a lexically reworded rubric (\textit{Paraphrased}). These variants preserve the intended scoring criteria but change the surface form or ordering of the rubric. Examples of the rubric transformations are provided in Appendix~\ref{sec:appendix_rubrics}. 

\begin{figure*}[t]
\centering
\includegraphics[width=\textwidth]{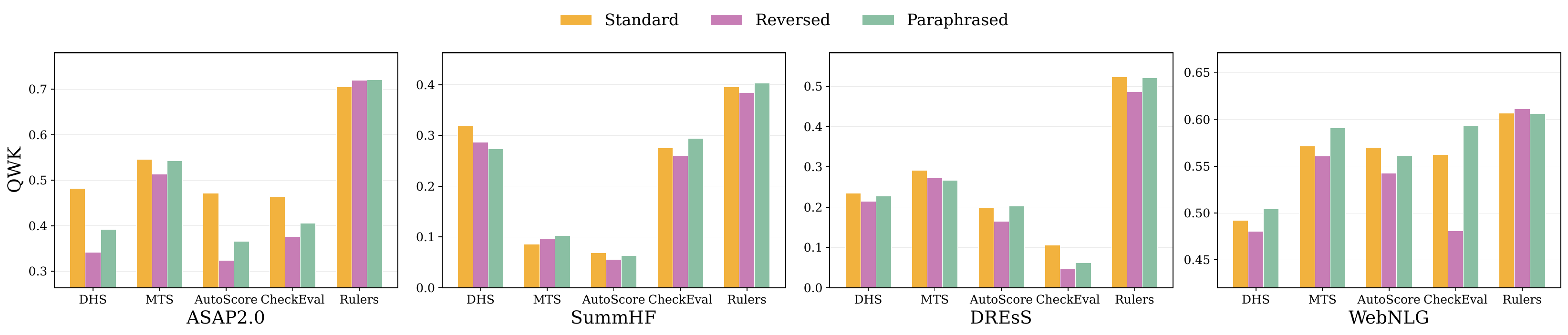}
\caption{
Rubric sensitivity analysis using \texttt{GPT-4o-mini}. We compare QWK under three semantically equivalent rubric presentations: Standard, Reversed, and Paraphrased.
}
\label{fig:robustness}
\end{figure*}

The perturbation results show that \textsc{Rulers} is generally robust to changes in rubric wording and ordering. Its QWK remains relatively stable across the three variants, suggesting that translating the rubric into a locked scoring specification and then applying post-hoc distribution alignment can reduce sensitivity to superficial prompt changes. A similar, though weaker, stabilizing effect can be observed for methods with explicit score-level post-processing, such as MTS, because the final scores are adjusted after the raw LLM judgments are produced. In contrast, methods that rely more directly on the raw rubric presentation tend to show larger shifts when the rubric is reordered or paraphrased. The robustness gap is smaller for settings with simple, low-dimensional rubrics, such as WebNLG, where the original scoring criteria are already concise and less ambiguous.

\subsection{Component Ablation Study}
\label{subsec:ablation}

We ablate the three phases of \textsc{Rulers} using the \texttt{GPT-4o-mini} backbone. The \textit{w/o Phase I locking} variant replaces the locked bundle with runtime raw-rubric interpretation; \textit{w/o Phase II evidence} keeps checklist execution but removes extractive evidence grounding and verification; and \textit{w/o Phase III distribution alignment} keeps the structured scorer and ridge-based latent mapping but removes the monotone mapping to the empirical human score distribution. Table~\ref{tab:ablation} reports QWK as bootstrap mean$\pm$SE.

\providecommand{\tightpm}{\!\pm\!}

\begin{table*}[t]
\centering
\small
\setlength{\tabcolsep}{6pt}
\renewcommand{\arraystretch}{1.10}
\caption{
Component ablation study of \textsc{Rulers} using \texttt{GPT-4o-mini}. Values are QWK reported as mean$\pm$bootstrap SE. Each ablation removes one phase while retaining the remaining pipeline as much as possible.
}
\label{tab:ablation}
\resizebox{\textwidth}{!}{
\begin{tabular}{lcccc}
\toprule
\textbf{Setting} 
& \textbf{ASAP 2.0} 
& \textbf{SummHF} 
& \textbf{DREsS} 
& \textbf{WebNLG} \\
\midrule
\textbf{Full \textsc{Rulers}}
& $\mathbf{0.7077\tightpm0.0058}$
& $\mathbf{0.3984\tightpm0.0305}$
& $\mathbf{0.5292\tightpm0.0203}$
& $\mathbf{0.6135\tightpm0.0294}$ \\

w/o Phase I locking
& $0.6766\tightpm0.0453$
& $0.3405\tightpm0.0305$
& $0.5055\tightpm0.0231$
& $0.5476\tightpm0.0409$ \\

w/o Phase II evidence
& $0.6833\tightpm0.0408$
& $0.3737\tightpm0.0260$
& $0.5088\tightpm0.0218$
& $0.5613\tightpm0.0319$ \\

w/o Phase III distribution alignment
& $0.6569\tightpm0.0601$
& $0.3404\tightpm0.0207$
& $0.4912\tightpm0.0230$
& $0.5040\tightpm0.0189$ \\
\bottomrule
\end{tabular}
}
\vspace{-4pt}
\end{table*}

The ablation results indicate that all three phases contribute to the final scoring performance. Removing Phase I weakens consistency because the model must reinterpret the rubric during scoring. Removing Phase II reduces auditability because scores are no longer tied to verifiable textual support. Removing Phase III distribution alignment makes the structured signals less matched to human score boundaries, even though the ridge-based latent mapping is used. These components also affect properties beyond QWK: Phase I supports consistent instance-level comparison through a fixed scoring specification, Phase II improves interpretability through checkable evidence, and Phase III reduces severity or leniency mismatch with human raters. The full framework is therefore strongest when rubric stability, evidence-grounded attribution, and human-scale alignment are combined.

\section{Related Work}
\label{sec:related}

\paragraph{Rubric-Based Text Evaluation with LLM Judges.}
Open-ended text evaluation has long relied on human-defined criteria because reference-based metrics such as BLEU and ROUGE often fail to capture higher-level qualities, including factual consistency, discourse organization, and instruction fulfillment \citep{papineni2002bleu,lin2004rouge,bhandari2020reevaluating,fabbri2021summeval}. Recent LLM-as-a-judge methods provide a more flexible alternative by using large language models as scalable evaluators, typically conditioning the judge on the input, evaluation criteria, and task-specific scoring instructions \citep{zheng2023judging,liu2023geval,gu2026survey}. This paradigm supports evaluation of complex and subjective qualities that are difficult to express through fixed lexical metrics, and has been applied across a growing range of generation and assessment settings. Large-scale comparisons further suggest that LLM judges can approximate human judgments in some tasks, although their reliability varies substantially across evaluator models, datasets, and evaluated properties \citep{bavaresco2025llms}. At the same time, a growing body of work has documented systematic sources of instability and bias in LLM-based evaluation. Pairwise judgments can change when the order of candidate responses is reversed \citep{wang2024fair}, while semantically equivalent evaluation instructions can produce noticeably different preferences \citep{zhou2024fairer}. Broader studies have identified multiple cognitive biases in LLM evaluators and gaps between model and human preferences \citep{koo2024benchmarking}. Together, these findings suggest that strong language-generation ability does not automatically translate into stable or human-aligned evaluation behavior. Our work focuses on the more specific setting of rubric-governed text scoring, where an explicit human-authored rubric and human score scale define the target standard, and studies how this scoring intent can be transferred reliably to a frozen LLM judge.

\paragraph{Structured, Evidence-Grounded, and Calibrated Evaluation.}
Prior work has addressed different aspects of this reliability problem by introducing additional structure into the evaluation process. Fine-grained evaluator models and protocols, such as Prometheus, FLASK, and HD-Eval, organize evaluation around customized rubrics, skill dimensions, or hierarchical criteria, allowing complex quality judgments to be decomposed into more specific components \citep{kim2024prometheus,ye2024flask,liu2024hdeval}. Multi-trait scoring follows a related principle by evaluating individual rubric dimensions separately before combining them into an overall score \citep{lee2024unleashing}. Checklist-based and component-based approaches further reduce reliance on unconstrained holistic judgments by converting broad criteria into simpler intermediate decisions or by extracting rubric-relevant scoring components before aggregation \citep{lee2025checkeval,wang2026autoscore}. Beyond decomposition, complementary approaches have explored multi-judge aggregation, entailment-based verification, and explicit rule-based evaluation and aggregation to improve the reliability of model-based judgments \citep{verga2024juries,laban2022summac,li2024pedants}. A separate line of work studies calibration, using human-scored examples to reduce mismatches between model outputs and human scoring behavior \citep{liu2024autocalibrate}. Collectively, these studies show that decomposition, structured intermediate judgments, verification, aggregation, and calibration can each address different weaknesses of automated evaluation. \textsc{Rulers} builds on these directions but studies them jointly within a criteria-transfer protocol for frozen, rubric-governed LLM judges.

\section{Conclusion}

This paper studies reliable rubric-based text evaluation with frozen LLM judges. Rather than viewing rubric-based scoring as a one-step prompting problem, we formulate it as a criteria-transfer problem: the goal is to translate human rubric intent into a scoring protocol that can be applied consistently, inspected through supporting evidence, and interpreted on the same score scale used by human raters. This perspective highlights three recurring sources of failure in LLM-based judging---rubric execution drift, unverifiable score attribution, and human-scale misalignment---and treats them as connected parts of a single evaluation problem.

To address these challenges, we introduce \textsc{Rulers}, a three-stage inference-time framework that first converts a source rubric into a reusable locked specification, then executes that specification through evidence-grounded structured scoring, and finally aligns the resulting model signals with human score boundaries through post-hoc calibration. Across four rubric-governed benchmarks and multiple frozen backbone models, \textsc{Rulers} achieves stronger agreement with human scores in most evaluated settings. Its predictions also more closely follow empirical human score distributions and remain more stable under semantically equivalent changes to rubric wording and criterion order. Although the improvement is not uniform for every dataset and backbone, the overall results show that reliable rubric-based evaluation depends on more than model capability or a carefully written prompt.

Our additional analyses help clarify where these gains come from. Applying the same calibration procedure to baseline outputs does not consistently reproduce the performance of \textsc{Rulers}, indicating that post-hoc score alignment alone is insufficient when the upstream scoring signals are weak or poorly structured. The component ablations further show that removing rubric locking, evidence grounding, or distribution alignment each reduces performance, supporting the view that the three stages address complementary parts of the scoring problem. Together, these results suggest that dependable LLM judging benefits from fixing the interpretation of evaluation criteria, connecting judgments to traceable textual support, and explicitly aligning model-derived signals with human scoring conventions. More broadly, our findings encourage treating an LLM judge not simply as a model that generates a score, but as one component of a controlled scoring protocol built around the human evaluation standard. Future work can extend this framework to additional evaluation domains, study transfer across related rubrics and score scales, and develop stronger evidence-grounding mechanisms for criteria that depend on holistic or implicit properties of a text.

\section{Acknowledgement}
The work was partially supported by Cisco Faculty Award, NSF award \#2442477, \#2550203, \#2616632, \#2623317, \#2536297 and \#2613637. The views and conclusions in this paper should not be interpreted as representing any funding agencies.

\section*{Limitations}

Our study focuses on rubric-based evaluation with a frozen, black-box LLM judge. This setting makes \textsc{Rulers} applicable to off-the-shelf models without fine-tuning, but it also introduces several practical limitations.

\paragraph{Dependence on the source rubric.}
\textsc{Rulers} transfers a human-authored rubric into a locked task-level bundle. This process improves consistency after the bundle is constructed, but it still depends on the quality of the original rubric. If the source rubric is vague, incomplete, or misaligned with the intended scoring standard, the induced bundle may preserve these ambiguities. Thus, \textsc{Rulers} is best viewed as a framework for stabilizing and operationalizing a given rubric, rather than as a method for automatically correcting flawed rubric design.

\paragraph{Frozen black-box judge setting.}
Our framework does not update the underlying LLM judge. This design is useful when only API access is available, but it also means that \textsc{Rulers} cannot directly change the model's internal reasoning ability or domain knowledge. The final scoring behavior therefore still depends partly on the capability and reliability of the chosen backbone model.

\paragraph{Calibration data requirement.}
The human-scale alignment stage requires a labeled calibration set. Although our experiments show that moderate calibration sizes can already be effective, calibration still depends on the availability and quality of human-scored examples. When a task has very few labeled examples, noisy labels, or a substantially different score distribution, recalibration may be necessary before applying the framework reliably.

\paragraph{Generalization to new rubrics and domains.}
\textsc{Rulers} is designed as a task-level protocol: each new evaluation setting requires constructing a locked rubric bundle and fitting the calibration mapping for the corresponding score scale. While this design supports stable execution within a task, applying the framework to a substantially different domain, rubric format, or scoring convention may require re-specification and recalibration. Future work can further study how rubric bundles and calibration mappings transfer across related tasks.

\section*{Ethical Considerations}

\textsc{Rulers} is designed to improve the reliability of LLM-based rubric evaluation by fixing task-level criteria, grounding decisions in structured evidence, and aligning model-derived signals to human score scales. Nevertheless, automated scoring systems should be used carefully, especially when evaluation results may affect individuals or institutions.

\paragraph{Appropriate use and human oversight.}
Even with improved stability and auditability, \textsc{Rulers} should not be treated as a complete replacement for expert judgment in high-stakes settings. Automated scores may still reflect rubric ambiguity, calibration noise, or model-specific biases. We recommend using \textsc{Rulers} as a decision-support tool and maintaining human review when evaluation outcomes have substantial consequences.

\paragraph{Privacy and data handling.}
Rubric-based evaluation may involve sensitive text, such as student writing or user-generated responses. When black-box LLM services are used, input text may be sent to external providers. Practitioners should minimize unnecessary data exposure, redact personally identifiable information when possible, and follow applicable privacy and data-governance requirements.

\paragraph{Fairness and accountability.}
LLM-based judges may behave differently across topics, writing styles, language backgrounds, or demographic groups. Calibration can improve agreement with a reference distribution, but it does not by itself guarantee fairness across all subgroups. Deployments should include subgroup-level auditing, documentation of rubric versions and calibration data, and clear procedures for human appeal or review.

\paragraph{Risks of gaming and misuse.}
A transparent and standardized evaluator may be easier to optimize against if users learn its scoring patterns. This risk is not unique to \textsc{Rulers}, but it is important for any rubric-based automated evaluation system. Regular rubric review, anomaly detection, and human spot checks can help reduce gaming and unintended over-optimization.

\paragraph{Disclosure of AI assistance.}
Generative AI tools were used to assist with language refinement and implementation prototyping during the development of this work. All authors remain fully responsible for the correctness, originality, and integrity of the methods, experiments, results, and writing.

\clearpage
\bibliography{ARR}

\clearpage
\appendix
\section{Appendix}
\label{app:appendix}


\subsection{Checklist Item Types and Evidence Operators}
\label{app:evidence_operator_types}

Table~\ref{tab:evidence_operator_types} summarizes the evidence-grounding types used in \textsc{Rulers}. Each checklist item is assigned an evidence-related type during Phase I bundle construction, and the corresponding evidence operator determines what form of support the judge should provide during Phase II execution. This design allows the same locked rubric bundle to support local sentence-level evidence, paragraph-level support, document-level diagnostics, and weakly groundable criteria while preserving a unified structured output for downstream calibration.

\subsection{Dataset Details}
\label{sec:appendix_datasets}

We use all datasets and model APIs in accordance with their licenses, terms of use, and research-use conditions. Table~\ref{tab:dataset_stats} summarizes the key statistics of the evaluation benchmarks used in this study. We report the number of held-out test samples, the score range, and the label distribution statistics measured by mean $\mu$ and standard deviation $\sigma$. For ASAP 2.0, SummHF, and DREsS, the target labels are the provided human scores. For WebNLG, the original human evaluation does not provide a single holistic score; instead, it reports trait-level human scores for semantics, grammar, and fluency. We therefore construct the target score by summing the three trait scores, resulting in a 3--9 score range.

\subsection{Baseline Protocol Comparison}
\label{app:baseline_protocol}

Table~\ref{tab:baseline_protocol} summarizes the protocol-level differences among the baselines and \textsc{Rulers}. The comparison focuses on how each method handles the rubric, obtains LLM judgments, uses evidence, and converts raw outputs into final scores.

\subsection{Average Token Usage}
\label{app:token_usage}

Table~\ref{tab:token_usage} reports the average total tokens used per evaluated sample with \texttt{GPT-4o-mini}. The values include both prompt-side and completion-side tokens across all LLM calls required by each method. Methods with multiple LLM calls per sample, such as MTS, naturally incur higher token usage, while direct scoring methods are more lightweight.

\subsection{Rubric Transformation Examples}
\label{sec:appendix_rubrics}

To test robustness to rubric presentation, we construct three semantically equivalent rubric variants for each dataset: the original rubric, an order-reversed rubric, and a paraphrased rubric. Table~\ref{tab:rubric_comparison} illustrates the construction using a generic ``Clarity'' criterion. The \textit{Standard} and \textit{Reversed} variants use identical text but differ in presentation order, while the \textit{Paraphrased} variant preserves the scoring intent with different surface wording.

\subsection{Calibration-Gain Diagnostic}
\label{app:calibration_gain_diagnostic}

To further separate the effect of calibration from the upstream structured scoring stages, we compare the QWK gain obtained by applying the same Phase-III alignment to DHS with the remaining gap between calibrated DHS and \textsc{Rulers}. Table~\ref{tab:calibration_gain_diagnostic} summarizes this comparison using $N=200$ calibration examples.

\begin{table}[t]
\centering
\scriptsize
\setlength{\tabcolsep}{4pt}
\renewcommand{\arraystretch}{1.10}
\caption{
Calibration-gain diagnostic using \texttt{GPT-4o-mini} with $N=200$. Values report QWK differences.
}
\label{tab:calibration_gain_diagnostic}
\resizebox{\columnwidth}{!}{
\begin{tabular}{lcc}
\toprule
\textbf{Dataset}
& \textbf{DHS $\rightarrow$ DHS+Calib.}
& \textbf{DHS+Calib. $\rightarrow$ \textsc{Rulers}} \\
\midrule

ASAP 2.0
& $+0.0829$
& $+0.1664$ \\

SummHF
& $+0.0380$
& $+0.0329$ \\

DREsS
& $+0.1053$
& $+0.1844$ \\

WebNLG
& $+0.0684$
& $+0.0527$ \\

\bottomrule
\end{tabular}
}
\end{table}

Calibration improves the unstructured DHS baseline on all four datasets, but a positive gap to \textsc{Rulers} remains in every case. This provides a lightweight diagnostic that the gains of \textsc{Rulers} are not explained by calibration alone and also depend on the structured signals produced by the preceding stages.

\subsection{Full Calibration-Control Results}
\label{app:calib_control_full}

Table~\ref{tab:calib_control_full} reports the full calibration-control results for all methods under calibration sizes $N\in\{50,100,150,200\}$. For baselines, we apply the same second-order ridge plus monotone distribution-alignment procedure to their final predicted scores. For \textsc{Rulers}, the same procedure is applied to its structured scoring signals. All mappings are fitted only on the calibration split and then fixed for held-out evaluation.

\subsection{Rubric-Transfer Fidelity}
\label{app:rubric_transfer_fidelity}

We further examine whether the Phase-I rubric-transfer process faithfully operationalizes the source rubric rather than introducing unsupported scoring criteria. For each dataset, we repeat the transformation three times using \texttt{GPT-4o} and summarize the induced traits, checklist composition, evidence types, and unsupported items in Table~\ref{tab:rubric_transfer_sanity}. Across datasets, the intended rubric structure is consistently preserved, with few or no unsupported checklist items. These results suggest that Phase I largely preserves the source scoring intent while converting the original rubric into a structured and executable specification.


\FloatBarrier

\begin{table*}[t]
\centering
\small
\setlength{\tabcolsep}{4pt}
\renewcommand{\arraystretch}{1.12}
\caption{
Checklist item types and evidence operators. The type is assigned during Phase I bundle construction and determines the required evidence form during Phase II execution.
}
\label{tab:evidence_operator_types}
\resizebox{\textwidth}{!}{
\begin{tabular}{p{0.18\textwidth}p{0.24\textwidth}p{0.25\textwidth}p{0.30\textwidth}}
\toprule
\textbf{Type} 
& \textbf{Applicable criteria} 
& \textbf{Evidence operator} 
& \textbf{Verification / use} \\
\midrule

\texttt{local\_quote}
& Claim, example, grammar error, specific factual or lexical cue
& Sentence-level quote
& Verify that the quoted text is grounded in the cited sentence. \\

\texttt{span\_level}
& Coherence, argument development, organization, progression of ideas
& Paragraph-level span with supporting sentence quote
& Verify that the cited support is grounded in the corresponding paragraph or paragraph-level unit. \\

\texttt{global\_diagnostic}
& Tone, creativity, overall flow, holistic quality pattern
& Document-level diagnostic with supporting spans
& Use a compact diagnostic statement together with representative supporting sentence or paragraph spans. \\

\texttt{weakly\_groundable}
& Subjective preference, implicit quality judgment, criteria that cannot be fully justified by a short quote
& Lower-confidence decision or human-review flag
& Retain the checklist decision but increase uncertainty or mark the instance for human review when evidence is not directly observable. \\

\bottomrule
\end{tabular}
}
\end{table*}

\begin{table*}[t]
\centering
\small
\setlength{\tabcolsep}{6pt}
\renewcommand{\arraystretch}{1.10}
\caption{
Dataset statistics of the evaluation benchmarks. Label statistics are reported as mean and standard deviation of human target scores on the held-out test split. For WebNLG, the target score is constructed by summing the semantics, grammar, and fluency scores.
}
\label{tab:dataset_stats}
\resizebox{\textwidth}{!}{
\begin{tabular}{l l c c c}
\toprule
\textbf{Dataset} 
& \textbf{Task Type} 
& \textbf{Test Size} 
& \textbf{Score Range} 
& \textbf{Label Stats ($\mu \pm \sigma$)} \\
\midrule
\textbf{ASAP 2.0} 
& Arg. Essay 
& 7,421 
& 1 -- 6 
& 2.92 $\pm$ 1.01 \\

\textbf{SummHF}   
& Summarization 
& 1,038 
& 1 -- 7 
& 4.69 $\pm$ 1.74 \\

\textbf{DREsS}    
& EFL Essay 
& 1,979 
& 3 -- 15 
& 9.94 $\pm$ 2.51 \\

\textbf{WebNLG}    
& Data-to-text 
& 611 
& 3 -- 9 
& 7.03 $\pm$ 1.76 \\
\bottomrule
\end{tabular}
}
\end{table*}

\begin{table*}[t]
\centering
\small
\setlength{\tabcolsep}{4pt}
\renewcommand{\arraystretch}{1.12}
\caption{Protocol-level comparison of baselines and \textsc{Rulers}.}
\label{tab:baseline_protocol}
\resizebox{\textwidth}{!}{
\begin{tabular}{p{0.13\textwidth}p{0.21\textwidth}p{0.22\textwidth}p{0.22\textwidth}p{0.18\textwidth}}
\toprule
\textbf{Method} 
& \textbf{Rubric handling} 
& \textbf{LLM scoring} 
& \textbf{Evidence / quote use} 
& \textbf{Final score} \\
\midrule

DHS &
Raw rubric used directly in one prompt &
Produces one holistic score directly &
May include rationale; no required quote grounding &
Direct parsed score \\

MTS &
Rubric decomposed into trait-level criteria &
Scores each trait separately &
No required extractive evidence or quote verification &
Aggregate trait scores \\

AutoScore &
Rubric guides component recognition or evidence extraction &
Uses extracted components before final scoring &
Uses rubric-relevant extracted components; no verbatim quote verifier &
Score from scoring agent \\

CheckEval &
Criteria converted into checklist questions &
Answers checklist questions with constrained responses &
Traceable checklist decisions; no verbatim quote verifier &
Aggregate checklist responses \\

\textsc{Rulers} &
Rubric induced and locked as bundle $\mathcal{B}$ &
Executes locked checklist with 0/1/2 decisions and trait scores &
Requires structured evidence objects with verifiable quotes when applicable &
Ridge regression + quantile mapping \\
\bottomrule
\end{tabular}
}
\end{table*}

\begin{table*}[t]
\centering
\small
\setlength{\tabcolsep}{6pt}
\renewcommand{\arraystretch}{1.10}
\caption{Average total token usage per evaluated sample using \texttt{GPT-4o-mini}.}
\label{tab:token_usage}
\resizebox{\textwidth}{!}{
\begin{tabular}{lccccc}
\toprule
\textbf{Dataset}
& \textbf{DHS}
& \textbf{MTS}
& \textbf{AutoScore}
& \textbf{CheckEval}
& \textbf{\textsc{Rulers}} \\
\midrule
ASAP 2.0
& 1421.21
& 5774.13
& 3720.40
& 1254.52
& 4811.96 \\

SummHF
& 1532.79
& 3843.38
& 2155.08
& 1026.08
& 3440.15 \\

DREsS
& 1048.09
& 3936.28
& 2218.14
& 1070.67
& 3591.84 \\

WebNLG
& 485.86
& 2104.11
& 1347.55
& 752.20
& 1868.87 \\
\bottomrule
\end{tabular}
}
\end{table*}

\begin{table*}[t]
\centering
\scriptsize
\setlength{\tabcolsep}{3.5pt}
\renewcommand{\arraystretch}{1.10}
\caption{
Phase-I rubric-transfer fidelity across three independent runs. Evidence-type counts correspond to local quote (Local), span level (Span), global diagnostic (Global), and weakly groundable (Weak). Numeric values are mean$\pm$SD.
}
\label{tab:rubric_transfer_sanity}
\resizebox{\textwidth}{!}{
\begin{tabular}{l p{0.19\textwidth} p{0.25\textwidth} c c c c c c}
\toprule
\textbf{Dataset}
& \textbf{Source-rubric traits}
& \textbf{Induced traits}
& \textbf{Items}
& \textbf{Local}
& \textbf{Span}
& \textbf{Global}
& \textbf{Weak}
& \textbf{Unsupported} \\
\midrule

ASAP 2.0
& Holistic rubric; no explicit trait-level ratings
& ClaimPosition; EvidenceElaboration; OrganizationCoherence; LanguageConventions
& $20.00\pm0.00$
& $6.00\pm1.73$
& $7.00\pm0.00$
& $7.00\pm1.73$
& $0.00\pm0.00$
& $0.33\pm0.58$ \\

SummHF
& Coherence; Accuracy; Coverage
& Same as source
& $20.00\pm0.00$
& $11.67\pm7.23$
& $4.67\pm4.04$
& $3.67\pm3.21$
& $0.00\pm0.00$
& $0.00\pm0.00$ \\

DREsS
& Content; Organization; Language
& Same as source
& $20.00\pm0.00$
& $8.33\pm1.53$
& $7.00\pm0.00$
& $4.67\pm1.53$
& $0.00\pm0.00$
& $0.00\pm0.00$ \\

WebNLG
& Semantics; Grammar; Fluency
& Same as source
& $20.00\pm0.00$
& $10.67\pm8.08$
& $4.00\pm3.46$
& $3.33\pm2.89$
& $2.00\pm1.73$
& $0.00\pm0.00$ \\

\bottomrule
\end{tabular}
}
\end{table*}

\begin{table*}[t]
\centering
\scriptsize
\setlength{\tabcolsep}{4pt}
\renewcommand{\arraystretch}{1.08}
\caption{
Full calibration-control results across calibration sizes. Values are QWK mean$\pm$bootstrap SE.
}
\label{tab:calib_control_full}
\resizebox{\textwidth}{!}{
\begin{tabular}{llcccc}
\toprule
\textbf{Dataset}
& \textbf{Method}
& \textbf{$N=50$}
& \textbf{$N=100$}
& \textbf{$N=150$}
& \textbf{$N=200$} \\
\midrule

\multirow{5}{*}{ASAP 2.0}
& DHS
& $0.5413\tightpm0.0079$
& $0.5414\tightpm0.0080$
& $0.5412\tightpm0.0081$
& $0.5413\tightpm0.0081$ \\
& MTS
& $0.4956\tightpm0.0082$
& $0.4954\tightpm0.0081$
& $0.5299\tightpm0.0083$
& $0.4955\tightpm0.0080$ \\
& AutoScore
& $0.4399\tightpm0.0081$
& $0.4928\tightpm0.0086$
& $0.4701\tightpm0.0084$
& $0.4701\tightpm0.0084$ \\
& CheckEval
& $0.5457\tightpm0.0081$
& $0.5450\tightpm0.0082$
& $0.5457\tightpm0.0082$
& $0.5457\tightpm0.0082$ \\
& \textsc{Rulers}
& $\mathbf{0.6929\tightpm0.0060}$
& $\mathbf{0.7024\tightpm0.0058}$
& $\mathbf{0.7074\tightpm0.0061}$
& $\mathbf{0.7077\tightpm0.0058}$ \\

\midrule

\multirow{5}{*}{SummHF}
& DHS
& $\mathbf{0.3618\tightpm0.0267}$
& $\mathbf{0.3743\tightpm0.0254}$
& $0.3738\tightpm0.0256$
& $0.3655\tightpm0.0260$ \\
& MTS
& $0.0621\tightpm0.0230$
& $0.0905\tightpm0.0290$
& $0.0896\tightpm0.0295$
& $0.0900\tightpm0.0294$ \\
& AutoScore
& $0.0658\tightpm0.0174$
& $0.0656\tightpm0.0173$
& $0.0656\tightpm0.0172$
& $0.0658\tightpm0.0172$ \\
& CheckEval
& $0.3244\tightpm0.0267$
& $0.3683\tightpm0.0269$
& $0.3705\tightpm0.0258$
& $0.3891\tightpm0.0269$ \\
& \textsc{Rulers}
& $0.2814\tightpm0.0302$
& $0.3388\tightpm0.0308$
& $\mathbf{0.3797\tightpm0.0301}$
& $\mathbf{0.3984\tightpm0.0305}$ \\

\midrule

\multirow{5}{*}{DREsS}
& DHS
& $0.2862\tightpm0.0182$
& $0.3439\tightpm0.0215$
& $0.3453\tightpm0.0213$
& $0.3448\tightpm0.0216$ \\
& MTS
& $0.3467\tightpm0.0210$
& $0.2988\tightpm0.0195$
& $0.3379\tightpm0.0205$
& $0.3380\tightpm0.0207$ \\
& AutoScore
& $0.2465\tightpm0.0229$
& $0.2341\tightpm0.0219$
& $0.2369\tightpm0.0217$
& $0.2436\tightpm0.0231$ \\
& CheckEval
& $0.3123\tightpm0.0206$
& $0.3293\tightpm0.0219$
& $0.3118\tightpm0.0207$
& $0.3284\tightpm0.0217$ \\
& \textsc{Rulers}
& $\mathbf{0.4624\tightpm0.0200}$
& $\mathbf{0.4977\tightpm0.0204}$
& $\mathbf{0.5186\tightpm0.0207}$
& $\mathbf{0.5292\tightpm0.0203}$ \\

\midrule

\multirow{5}{*}{WebNLG}
& DHS
& $0.5404\tightpm0.0209$
& $0.5605\tightpm0.0211$
& $0.5607\tightpm0.0205$
& $0.5608\tightpm0.0209$ \\
& MTS
& $0.5660\tightpm0.0256$
& $0.5781\tightpm0.0262$
& $0.5747\tightpm0.0258$
& $0.5738\tightpm0.0266$ \\
& AutoScore
& $0.5593\tightpm0.0248$
& $0.5633\tightpm0.0236$
& $0.5650\tightpm0.0239$
& $0.5651\tightpm0.0237$ \\
& CheckEval
& $\mathbf{0.6057\tightpm0.0250}$
& $0.5880\tightpm0.0251$
& $0.5821\tightpm0.0252$
& $0.5822\tightpm0.0250$ \\
& \textsc{Rulers}
& $0.5823\tightpm0.0254$
& $\mathbf{0.5997\tightpm0.0251}$
& $\mathbf{0.6002\tightpm0.0263}$
& $\mathbf{0.6135\tightpm0.0294}$ \\
\bottomrule
\end{tabular}
}
\end{table*}

\begin{table*}[t]
\centering
\small
\setlength{\tabcolsep}{4pt}
\renewcommand{\arraystretch}{1.08}
\caption{
Comparison of the three rubric variants used for robustness testing. The \textit{Reversed} variant uses identical text to the \textit{Standard} one but inverts the sequential order in the prompt.
}
\label{tab:rubric_comparison}
\resizebox{\textwidth}{!}{
\begin{tabular}{p{0.12\textwidth}p{0.26\textwidth}p{0.26\textwidth}p{0.26\textwidth}}
\toprule
\textbf{Feature} 
& \textbf{Variant 1: Standard} 
& \textbf{Variant 2: Reversed} 
& \textbf{Variant 3: Paraphrased} \\
\midrule

\textbf{Order} 
& \textbf{Descending (5 $\to$ 1)} 
& \textbf{Ascending (1 $\to$ 5)} 
& \textbf{Descending (5 $\to$ 1)} \\

\midrule
\textbf{Score 5} &
The explanation is crystal clear and completely logical. &
The explanation is crystal clear and completely logical. &
The writing demonstrates distinct clarity and flawless reasoning. \\

\midrule
\textbf{Score 3} &
The explanation is understandable but has some confusing parts. &
The explanation is understandable but has some confusing parts. &
The writing is mostly comprehensible despite minor ambiguities. \\

\midrule
\textbf{Score 1} &
The explanation is confusing and impossible to follow. &
The explanation is confusing and impossible to follow. &
The writing is incoherent and completely lacks readability. \\
\bottomrule
\end{tabular}
}
\end{table*}

\clearpage

\subsection{Prompt Templates}
\label{sec:appendix_prompts}

To ensure reproducibility and model agnosticism, \textsc{Rulers} uses a unified prompting strategy across datasets. We categorize the prompts into two templates corresponding to the framework stages: (1) \textbf{Rubric Specification}, which converts the raw natural-language rubric into a locked, executable rubric bundle; and (2) \textbf{Evidence-Grounded Scoring}, which executes the locked bundle and requires structured, auditable evidence during inference. The templates below are simplified for readability while preserving the key evidence requirements used in our implementation.

\begin{jsoncasebox}{Phase I: Rubric Specification Template}
You are an expert writing-assessment designer.
You are a RUBRIC COMPILER, not a rubric author.
Output STRICT JSON matching the schema only.

We are scoring {INPUT_TYPE} on the official score scale
based on the OFFICIAL rubric below.

Input Context:
- Dataset Name: {DATASET_NAME}
- Raw Rubric: {RAW_RUBRIC_TEXT}

Your job is to compile the raw rubric into a FIXED
task-level rubric bundle.

Your Constraints:
1. Define EXACTLY {K} evaluation traits:
   {FIXED_TRAIT_LIST}

2. Create EXACTLY {J} checklist items total:
   - Checklist decisions use 0/1/2:
     0 = not present,
     1 = partially present,
     2 = clearly present.
   - Each item must be decisionable and auditable.

3. Evidence Requirements:
   - Each checklist item should be associated with
     evidence expectations whenever possible.
   - Evidence should be grounded in the evaluated text.
   - Evidence rules should specify whether evidence is
     expected at the sentence, paragraph, or document level.
   - Evidence quotes should be short, verbatim, and
     traceable to sentence or paragraph IDs.
   - Set evidence_rules.min_evidence_per_trait = {MIN_EV}.

4. Hard Constraints:
   - Must COVER all official key concepts.
   - Do NOT invent a new scoring system.
   - Remain faithful to the official rubric and score scale.

Output Format:
- Return a structured JSON object containing:
  traits, checklist items C01..C{J}, evidence rules,
  score boundaries, and bundle metadata.
- This output will be serialized, hashed, and locked.
\end{jsoncasebox}

\newpage

\begin{jsoncasebox}{Phase II: Evidence-Grounded Scoring Template}
You are a strict evidence-based judge.
Return JSON ONLY. Follow the schema exactly.
Do not provide long chain-of-thought. Use short,
auditable justifications.

Input Data:
- Locked Bundle Hash: {BUNDLE_HASH}
- Paragraph Bank: {PARAGRAPH_BANK}
- Sentence Bank: {SENTENCE_BANK}
- Locked Traits: {LOCKED_TRAITS_JSON}
- Locked Checklist: {LOCKED_CHECKLIST_JSON}
- Evidence Rules: {EVIDENCE_RULES_JSON}

Your Task:
1. Paragraph Outline:
   - For each paragraph, provide a short summary.
   - Select one topic_sent_id from the sentence bank.

2. Checklist Execution:
   - Rate ALL {J} checklist items exactly once.
   - Each decision must be one of {0, 1, 2}.
   - Decision=2 requires clear positive support.

3. Evidence Extraction:
   - For EACH trait, provide EXACTLY {MIN_EV}
     evidence quotes.
   - Each evidence item must include:
     sent_id and quote.
   - The quote must be a verbatim substring of the
     specified sentence text.
   - Evidence should support the corresponding trait
     score or checklist decisions.
   - If evidence is weak or incomplete, reflect this
     through lower confidence rather than inventing
     unsupported evidence.

4. Boundary Justification:
   - For each trait, provide a compact boundary check:
     explain why the score is not one level higher
     and not one level lower.

5. Anti-Halo Constraint:
   - Do NOT output identical scores across all traits
     unless equality is justified by boundary checks.
   - Avoid central tendency bias.

Return the structured evaluation object now.
\end{jsoncasebox}

\FloatBarrier
\clearpage
\section{Concrete ASAP 2.0 Case Study}
\label{app:case_study}

This appendix provides a shortened execution trace of \textsc{Rulers} on one ASAP 2.0 essay. The case is for illustration only. We show selected excerpts from the official rubric, the student response, the locked bundle, and the model output rather than the complete rubric, full essay, or full JSON artifact. The selected essay is de-identified as ASAP2-Case-01. It comes from the prompt ``A Cowboy Who Rode the Waves,'' where students write an argument from Luke's point of view to persuade others to participate in the Seagoing Cowboys program. The human holistic score is 4 on the 1--6 scale, and \textsc{Rulers} also assigns a final calibrated score of 4.

\subsection{Phase I Case: From Official Rubric to Locked Bundle}
\label{app:case_phase1}

\textsc{Rulers} uses the official holistic rating form as the source rubric. The full rubric defines a 1--6 scale and describes writing quality in terms of claim development, source-based evidence, organization, coherence, language use, and mechanics. Below we show selected clauses from the rubric before showing the locked bundle excerpt.

\begin{casebox}{Phase I-A: Official Holistic Rating Form, Selected Excerpts}
The official rubric asks raters to assign a holistic score from 1 to 6 after reading the essay. The distance between adjacent grades is treated as equal.

\smallskip
Score 6: The essay demonstrates clear and consistent mastery. It effectively and insightfully develops a point of view, uses clearly appropriate examples, reasons, and source-based evidence, is well organized and clearly focused, shows coherence and smooth progression, uses varied and accurate language, and is free of most errors.

\smallskip
Score 5: The essay demonstrates reasonably consistent mastery. It effectively develops a point of view, generally uses appropriate source-based evidence, is well organized and focused, and is generally free of most errors.

\smallskip
Score 4: The essay demonstrates adequate mastery, although it may have lapses. It develops a point of view, uses adequate examples or source-based evidence, is generally organized and focused, and may contain some language or mechanics errors.

\smallskip
Score 3: The essay demonstrates developing mastery. It may use inadequate evidence, show limited organization or focus, contain lapses in coherence, or include accumulated errors.

\smallskip
Score 2: The essay demonstrates little mastery. It may present a vague position, provide insufficient evidence, show poor organization, or contain frequent language errors.

\smallskip
Score 1: The essay demonstrates very little or no mastery. It may lack a viable point of view, provide little evidence, be disorganized or incoherent, or contain pervasive errors that interfere with meaning.
\end{casebox}

The Phase I output is a locked task-level bundle. The full bundle contains four traits, score anchors, 20 checklist items, evidence-grounding types, evidence operators, and evidence rules. The excerpt below shows only representative fields.

\begin{jsoncasebox}{Phase I-B: Abridged Locked Bundle Excerpt}
{
  "rubric_version": "v1",
  "bundle_hash": "[omitted]",

  "traits": {
    "ClaimPosition": {
      "6": "Clear, insightful, and consistent position.",
      "4": "Adequate position with competent reasoning.",
      "2": "Vague or seriously limited position."
    },

    "EvidenceElaboration": {
      "6": "Strong source-based evidence with explanation.",
      "4": "Adequate examples, reasons, or source evidence.",
      "2": "Insufficient or weakly connected evidence."
    },

    "OrganizationCoherence": {
      "6": "Clear focus and smooth progression.",
      "4": "Generally organized with some coherence.",
      "2": "Poor focus or serious coherence problems."
    },

    "LanguageConventions": {
      "6": "Varied language with few errors.",
      "4": "Generally appropriate language with some errors.",
      "2": "Frequent errors affecting clarity."
    }
  }
}
\end{jsoncasebox}

Each checklist item is assigned an evidence-grounding type and a corresponding evidence operator. This assignment is part of the locked bundle and is reused unchanged during scoring.

\begin{jsoncasebox}{Phase I-C: Representative Typed Checklist Items}
{
  "checklist_excerpt": [
    {
      "id": "C01",
      "dimension": "ClaimPosition",
      "criterion": "States a clear position responding to the prompt.",
      "decision_scale": "0=absent, 1=partial, 2=clear",
      "item_type": "local_quote",
      "evidence_operator": "sentence_quote"
    },
    {
      "id": "C07",
      "dimension": "EvidenceElaboration",
      "criterion": "Uses source-based evidence relevant to the claim.",
      "decision_scale": "0=absent, 1=partial, 2=clear",
      "item_type": "local_quote",
      "evidence_operator": "sentence_quote"
    },
    {
      "id": "C09",
      "dimension": "EvidenceElaboration",
      "criterion": "Explains how the evidence supports the position.",
      "decision_scale": "0=absent, 1=partial, 2=clear",
      "item_type": "span_level",
      "evidence_operator": "paragraph_span"
    },
    {
      "id": "C13",
      "dimension": "OrganizationCoherence",
      "criterion": "Orders reasons in a coherent progression.",
      "decision_scale": "0=absent, 1=partial, 2=clear",
      "item_type": "span_level",
      "evidence_operator": "paragraph_span"
    },
    {
      "id": "C16",
      "dimension": "OrganizationCoherence",
      "criterion": "Maintains an overall focused flow.",
      "decision_scale": "0=absent, 1=partial, 2=clear",
      "item_type": "global_diagnostic",
      "evidence_operator": "document_diagnostic_with_supporting_spans"
    },
    {
      "id": "C18",
      "dimension": "LanguageConventions",
      "criterion": "Controls grammar, spelling, and mechanics.",
      "decision_scale": "0=absent, 1=partial, 2=clear",
      "item_type": "local_quote",
      "evidence_operator": "sentence_quote"
    }
  ],

  "evidence_rules": {
    "quote_must_be_verbatim": true,
    "min_evidence_per_trait": 2,
    "typed_evidence": true
  }
}
\end{jsoncasebox}

The key point is that the holistic rubric is not used as free-form prompt context at every scoring step. It is first converted into a reusable specification containing trait anchors, checklist items, item-level evidence types, and evidence rules. This locked bundle is then reused for all ASAP 2.0 essays in the same run.

\subsection{Phase II Case: From Locked Bundle to Evidence-Grounded Output}
\label{app:case_phase2}

Before scoring, the essay is segmented into atomic units. Sentence-level units are used for local evidence, while paragraph-level units are available for span-level or document-level diagnostics. We do not reproduce the full student response. The excerpts below show selected units used by the judge. Spelling and grammar are preserved from the student response.

\begin{jsoncasebox}{Phase II-A: Selected Sentence and Paragraph Units}
{
  "sentences": [
    {
      "sent_id": 0,
      "para_id": 0,
      "text": "I think that you should defineatly join the Saegoing Cowboys, which is a group of people who go across the Atlantic ocean and the Pacific ocean to help people from World War ll."
    },
    {
      "sent_id": 6,
      "para_id": 1,
      "text": "In my opinion, I think people should join the Seagoing Cowboys program for a few different reasons."
    },
    {
      "sent_id": 14,
      "para_id": 2,
      "text": "\"It made me more aware of people of other countries and their needs.\""
    },
    {
      "sent_id": 19,
      "para_id": 3,
      "text": "My last reason that I think you should join the Seagoing Cowboys is that you will meet lots of people."
    },
    {
      "sent_id": 25,
      "para_id": 4,
      "text": "In conclusion, I strongly encourage you to join the Seagoing Cowboys."
    },
    {
      "sent_id": 28,
      "para_id": 4,
      "text": "This was an oppurtunity of a lifetime for me."
    }
  ],

  "paragraphs": [
    {
      "para_id": 1,
      "text": "In my opinion, I think people should join the Seagoing Cowboys program for a few different reasons. ..."
    },
    {
      "para_id": 2,
      "text": "One reason is that the program helps people understand other countries and their needs. \"It made me more aware of people of other countries and their needs.\" ..."
    },
    {
      "para_id": 4,
      "text": "In conclusion, I strongly encourage you to join the Seagoing Cowboys. This was an oppurtunity of a lifetime for me."
    }
  ]
}
\end{jsoncasebox}

Using the locked bundle and the segmented units, the judge fills the checklist and returns typed evidence. The next block shows representative outputs rather than all 20 checklist decisions.

\begin{jsoncasebox}{Phase II-B: Representative Typed Checklist Output}
{
  "essay_id": "ASAP2-Case-01",

  "checklist_ratings_excerpt": [
    {
      "id": "C01",
      "dimension": "ClaimPosition",
      "decision": 2,
      "item_type": "local_quote",
      "evidence_operator": "sentence_quote",
      "evidence": {
        "sent_id": 0,
        "quote": "I think that you should defineatly join the Saegoing Cowboys"
      }
    },
    {
      "id": "C07",
      "dimension": "EvidenceElaboration",
      "decision": 2,
      "item_type": "local_quote",
      "evidence_operator": "sentence_quote",
      "evidence": {
        "sent_id": 14,
        "quote": "It made me more aware of people of other countries and their needs."
      }
    },
    {
      "id": "C09",
      "dimension": "EvidenceElaboration",
      "decision": 1,
      "item_type": "span_level",
      "evidence_operator": "paragraph_span",
      "evidence": {
        "para_id": 2,
        "span": "One reason is that the program helps people understand other countries and their needs. \"It made me more aware of people of other countries and their needs.\""
      }
    },
    {
      "id": "C13",
      "dimension": "OrganizationCoherence",
      "decision": 1,
      "item_type": "span_level",
      "evidence_operator": "paragraph_span",
      "evidence": {
        "para_id": 1,
        "span": "In my opinion, I think people should join the Seagoing Cowboys program for a few different reasons."
      }
    },
    {
      "id": "C16",
      "dimension": "OrganizationCoherence",
      "decision": 1,
      "item_type": "global_diagnostic",
      "evidence_operator": "document_diagnostic_with_supporting_spans",
      "evidence": {
        "diagnostic": "The essay has a recognizable introduction, body reasons, and conclusion, but the progression is formulaic.",
        "supporting_spans": [
          {"sent_id": 6, "quote": "for a few different reasons"},
          {"sent_id": 25, "quote": "In conclusion, I strongly encourage you to join"}
        ]
      }
    },
    {
      "id": "C18",
      "dimension": "LanguageConventions",
      "decision": 1,
      "item_type": "local_quote",
      "evidence_operator": "sentence_quote",
      "evidence": {
        "sent_id": 28,
        "quote": "oppurtunity of a lifetime"
      }
    }
  ]
}
\end{jsoncasebox}

This typed evidence structure allows the same checklist execution format to handle different kinds of criteria. Local criteria are supported by sentence-level quotes, discourse-level criteria can use paragraph spans, and global criteria can return a compact diagnostic with supporting spans. If a criterion is weakly groundable, the output can retain the checklist decision while lowering confidence or marking the item for human review.

\begin{jsoncasebox}{Phase II-C: Weakly Groundable Indicator Example}
{
  "id": "C20",
  "dimension": "LanguageConventions",
  "criterion": "Overall readability is appropriate for the task.",
  "decision": 1,
  "item_type": "weakly_groundable",
  "evidence_operator": "lower_confidence_human_review",
  "evidence": {
    "diagnostic": "The essay remains understandable, but readability is affected by repeated spelling and wording errors.",
    "human_review_recommended": false,
    "confidence_adjustment": "lower"
  }
}
\end{jsoncasebox}

The verifier then applies the operator-specific grounding rule. Sentence-level quotes are checked against the cited sentence, paragraph-level spans are checked against the cited paragraph unit, and supporting spans for global diagnostics are checked against their corresponding units. Weakly groundable items are not treated as ordinary quote failures; instead, they contribute to uncertainty or human-review indicators.

\begin{jsoncasebox}{Phase II-D: Evidence Verification Summary}
{
  "sentence_quote_items": 4,
  "paragraph_span_items": 2,
  "global_diagnostic_items": 1,
  "weakly_groundable_items": 1,

  "raw_evidence_total": 8,
  "valid_evidence_total": 8,
  "invalid_evidence": 0,
  "human_review_recommended": false
}
\end{jsoncasebox}

The model also returns compact boundary checks for the trait scores. These are used as structured diagnostic signals, but they are not free-form rationales used to override the locked checklist.

\begin{jsoncasebox}{Phase II-E: Boundary Checks, Selected Excerpts}
{
  "ClaimPosition": 
    "The essay gives a clear recommendation, but the reasoning is not consistently insightful enough for a higher level.",

  "EvidenceElaboration": 
    "The essay uses source-based evidence, but some explanation remains general.",

  "OrganizationCoherence": 
    "The essay has an introduction and conclusion, but transitions are mechanical.",

  "LanguageConventions": 
    "Meaning is understandable, but spelling and mechanics errors are visible."
}
\end{jsoncasebox}

After checklist aggregation and evidence verification, \textsc{Rulers} produces a unified structured signal vector. Importantly, the downstream calibration interface remains the same: typed evidence affects the quality and reliability of upstream signals, but the calibration stage still receives trait scores and auxiliary indicators such as confidence, uncertainty, invalid evidence counts, and tie indicators.

\begin{jsoncasebox}{Phase II-F: Aggregate Structured Signals}
{
  "ClaimPosition": 4,
  "EvidenceElaboration": 4,
  "OrganizationCoherence": 3,
  "LanguageConventions": 3,

  "confidence": 0.85,
  "invalid_evidence": 0,
  "raw_evidence_total": 8,
  "valid_evidence_total": 8,
  "human_review_recommended": false,
  "tie": 0,
  "trait_range": 1
}
\end{jsoncasebox}

\subsection{Phase III Case: From Structured Signals to Final Score}
\label{app:case_phase3}

Phase III maps the structured signals to the human score scale. The evidence operators do not change the calibration model itself; instead, they affect the upstream structured signals that enter calibration. For this essay, the calibration feature vector is
\[
\phi_\theta(x;\mathcal{B})=[4,4,3,3,\log(1+2307),0.85,0.15,0],
\]
where the first four values are trait scores, followed by essay length, confidence, uncertainty, and the tie indicator. Since $\log(1+2307)\approx 7.74$, the feature vector is approximately
\[
[4,4,3,3,7.74,0.85,0.15,0].
\]

\begin{jsoncasebox}{Phase III: Calibration Trace}
{
  "dataset": "ASAP 2.0",
  "prompt": "A Cowboy Who Rode the Waves",
  "grade_level": 6,

  "human_score": 4,

  "raw_trait_scores": {
    "ClaimPosition": 4,
    "EvidenceElaboration": 4,
    "OrganizationCoherence": 3,
    "LanguageConventions": 3
  },

  "evidence_summary": {
    "sentence_quote_items": 4,
    "paragraph_span_items": 2,
    "global_diagnostic_items": 1,
    "weakly_groundable_items": 1,
    "invalid_evidence": 0,
    "human_review_recommended": false
  },

  "mean_raw_trait_score": 3.50,
  "final_score_cont": 4.00,
  "final_score": 4
}
\end{jsoncasebox}

This case illustrates the full \textsc{Rulers} path on ASAP 2.0 under typed evidence grounding. The official holistic rubric is transformed into a locked four-trait specification, each checklist item declares an evidence-grounding type, the frozen judge executes the specification with operator-specific evidence, and the resulting structured signals are calibrated to the human 1--6 score scale.

\end{document}